# Exploiting SIFT Descriptor for Rotation Invariant Convolutional Neural Network

Abhay Kumar
Voice Intelligence R&D
Samsung R&D Institute
Bangalore, India
abhay1.kumar@samsung.com

Nishant Jain
Voice Intelligence R&D
Samsung R&D Institute
Bangalore, India
nishant.jain@samsung.com

Chirag Singh
Voice Intelligence R&D
Samsung R&D Institute
Bangalore, India
c.singh@samsung.com

Suraj Tripathi
Voice Intelligence R&D
Samsung R&D Institute
Bangalore, India
suraj.tri@samsung.com

*Abstract*—This paper presents a novel approach to exploit the distinctive invariant features in convolutional neural network. The proposed CNN model uses Scale Invariant Feature Transform (SIFT) descriptor instead of the max-pooling layer. Max-pooling layer discards the pose, i.e., translational and rotational relationship between the low-level features, and hence unable to capture the spatial hierarchies between low and high level features. The SIFT descriptor layer captures the orientation and the spatial relationship of the features extracted by convolutional layer. The proposed SIFT Descriptor CNN therefore combines the feature extraction capabilities of CNN model and rotation invariance of SIFT descriptor. Experimental results on the MNIST and fashionMNIST datasets indicates reasonable improvements over conventional methods available in literature.

## I. Introduction

Convolutional Neural Network (CNN) has been widely used for many computer vision applications like classification [8], recognition etc. CNN models are able to learn complex feature representations from the images. Majority of the CNN architectures (e.g. Jarrett et al. [15]; Krizhevsky et al. [16]) have the alternating convolutional and max-pooling layers followed by fully connected network. Different activation functions [19] have been experimented for the abovementioned layers. A significant amount of research work to improve the basic CNN pipeline has been presented over the last few decades. This includes techniques such as feature pooling [9], dropout [11], Drop-Connect [12], batch normalization [17], deep CNN with multiple filters [18], improved weight initialization methods [6], deep residual learning [20], and data augmentation [22]. Most advances deal with building more sophisticated layers, applying effective regularization and improving weight initializations. However, the pooling layer has been equally important to make the CNN model invariance to data variation and perturbations. Graham et al [14] formulated the fractional max pooling to reduce the spatial size by non-integer factor. Lee et al [7] presented an approach to bring learning into the pooling operation by mixed or gated combination of average and max pooling operations and tree-structured fusion of learnable pooling filters. Zeiler et al [10] used stochastic pooling strategy to randomly pick the activation within each pooling region based on a multinomial distribution. Scherer et al [12] experimented with overlapping and non-overlapping pooling windows, and presented better performance of max pooling operation over subsampling operations. Yu et al [13] proposed Mixed pooling, which stochastically uses conventional max pooling and average pooling operations. Tolias et al [5] used the integral image for max-pooling operations on convolutional layer activations. Williams et al [4] introduced wavelet pooling to decompose features, generated by the convolutional layer, into a second level decomposition and discards the first-level subbands to reduce feature dimensions in a more structurally compact manner than pooling via neighborhood regions. It also addresses the overfitting problem encountered by max pooling.

Despite the abovementioned advances in conventional Convolutional Neural Network, there are still several shortcomings such as inability to generalize to novel viewpoints, and inability to capture the spatial relationship between the low-level features. These shortcomings link to fact that CNNs pay attention to the presence of certain key features on the image, ignoring their relative position as to each other. Convolutional neural networks have translation invariance as one major advantage. However, this invariance ignores how different features are interconnected with respect to each other. For instance, if we consider an image of a face, CNN will have difficulties capturing the relationship among mouth, eyes and nose features. Max pooling layers are mainly responsible for this effect. Because when we apply max pooling operations, we lose the precise locations of the mouth, eyes and nose and how the facial parts are related to each other.

Recently Hinton et al [1] and Sabour et al [2] proposed a novel Capsule Network (CapsNet) architecture to capture the hierarchical pose (translation and rotation) relationship. The activity vector of the capsule (group of neurons) encodes the instantiation parameters of an entity- an object or part of an object. CapsNet uses routing-by-agreement algorithm to preserve local relative position, which is discarded by max pooling. Capsule Networks strives for *translation equivariance* instead of *translation invariance*, allowing it to generalize to a greater degree from different viewpoints with less training data. Although the approach is quite promising, it has some shortcomings like slow learning process, crowding effect, and higher computational complexity.

In this paper, we propose a novel variation of CNN baseline model, where pooling layer is replaced by Scale

Invariant Feature Transform (SIFT) descriptor to preserve the spatial arrangement of low level features. We also experimented with hybrid max-SIFT variant of CNN model.

## II. PROPOSED ARCHITECTURE

The proposed model replaces the traditional max-pooling layer with SIFT feature based descriptor. SIFT descriptor helps retain the spatial orientations of the low-level features extracted by the convolutional layer. CNN is very good at learning the different kernel weights for feature extraction, but max pooling discards significant relationships among the extracted features. SIFT descriptor takes in account the histogram of the gradients in a given patch size, and thereby preserving the relative orientation of the extracted features.

### A. Motivation to use SIFT descriptor

Lowe [3] proposed a method to extract distinctive invariant image features that can be used to perform robust matching between different viewpoints of an object. SIFT features are scale and rotation invariant, and hence robust to substantial range of affine distortion, change in viewpoint, illumination and noise. Both spatial and frequency localization of the features reduces the effect of occlusion, clutter, or noise. High distinctiveness of SIFT features allows robust matching with just a single feature. The SIFT algorithm has two major stages- Keypoint Extraction and Keypoint Description. We only use the Keypoint description (refer Fig.1) stage in our proposed model architecture.
CNN is very good at extracting the distinct features (keypoints) for classification. We use the distinctiveness and orientation invariance properties of the SIFT features to overcome the limitations of max-pooling. The SIFT Descriptor CNN exploits both the feature extraction of CNN and local descriptor of the SIFT.

### B. SIFT descriptor Methodology

The Scale Invariant Feature Transform (SIFT) [3] keypoint descriptor gives 128 dimensional feature representation of the 16x16 neighborhood of keypoint. The SIFT descriptor methodology is explained for an image patch, $I(x,y)$ of size 16x16 around the keypoint. The SIFT descriptor consists of the following steps-

#### 1) Orientation Assignment

This step assigns orientation based on the local image gradient directions. This makes the descriptor rotation invariant as the keypoint descriptor is represented relative to the assigned orientation. For the given image patch, $I(x,y)$ the gradient magnitude, $m(x,y)$ and orientation, $\theta(x,y)$ are computed as shown below.

$$m(x,y) = \sqrt{(I(x+1,y) - I(x-1,y))^2 + (I(x,y+1) - I(x,y-1))^2} \quad (1)$$

$$\theta(x,y) = \operatorname{atan2}(I(x,y+1) - I(x,y-1), I(x+1,y) - I(x-1,y)) \quad (2)$$

The gradient magnitude and direction are calculated for every pixel in the given patch. The gradient magnitude are weighted by a Gaussian circular weighting function with $\sigma$, being one half of the descriptor window length. Gaussian blurring emphasizes the gradients that are closer to the center of the descriptor. An orientation histogram of the gradients orientations is computed for the patch. The histogram has 36 bins spanning 360 degree range of gradient orientations. The peak in the histogram signifies the dominant direction among the local gradients and acts as the reference direction for the given patch. Then, the local gradient distribution is transformed relative to this reference direction.

#### 2) Keypoint descriptor

This step computes a descriptor vector for each image patch such that the descriptor is highly distinctive and partially invariant to the remaining variations such as illumination, 3D viewpoint, etc.

The image patch (16x16) is divided into 4x4 subregions, where each subregion is 4x4 pixel neighborhood. The local gradients are already normalized with the reference direction in the previous step. An orientation histogram with 8 bins (each bin covering 45 degrees) is computed for each of the 16 subregions. The descriptor then becomes a vector of all the values of these histograms. Since there are 4×4=16 histograms, each having 8 bins, the descriptor vector has 128 elements. Then the vector is normalized to unit length; enhancing its invariance to affine changes in illumination. In order to reduce the effects of non-linear illumination, a threshold of 0.2 is applied and the vector is again normalized. The threshold of 0.2 was experimentally chosen [3] to reduce the effects of non-linear illumination.

### C. SIFT feature descriptor Implementation

The SIFT descriptor is part of the CNN network, so we need to make it differentiable for the back-propagation to update model weights. Hence, it needs to be implemented using the standard differentiable mathematical operations.

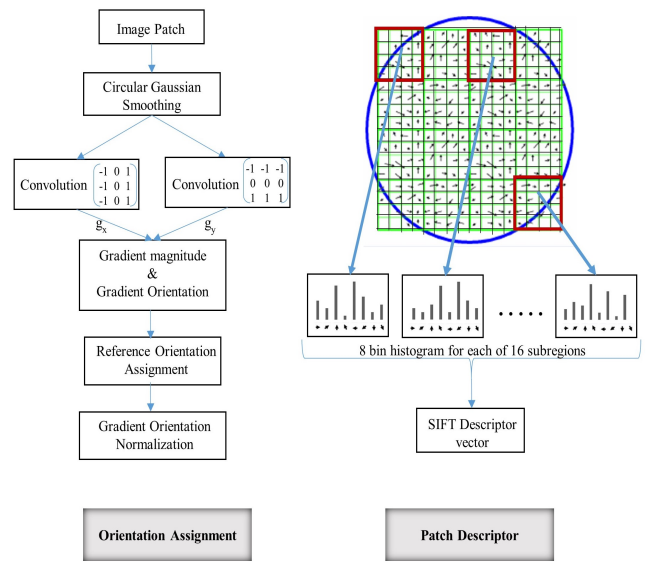

Fig. 1. High level block diagram for the SIFT Descriptor.

We implemented SIFT descriptor using both theano and tensorflow based operations. High level block diagram of the SIFT descriptor is shown in Fig.1. We have shown the SIFT descriptor methodology for image patch, size 16x16. We have extended the same methodology for patch size p x p (where p ≥ 16) for our experimental models. The only modification is dividing the patch (p x p) into 4x4 subregions, where each subregion is ( $p/4$ x $p/4$ ) pixel neighborhood in the keypoint descriptor step.

### III. DATASETS

We evaluated our proposed model against the following benchmark datasets.

#### A. MNIST

The MNIST [23] dataset (Modified National Institute of Standards and Technology database) is a large database of handwritten digits. The MNIST dataset comprises of 10-class handwritten digit. The MNIST dataset has 60,000 training images and 10,000 testing images. The black and white images from NIST were normalized to fit into a 28x28 pixel bounding box. This is commonly used for benchmarking machine learning algorithms.

#### B. fashionMNIST

Fashion-MNIST [24] is a dataset comprising of 28×28 grayscale images of 70,000 fashion products. This has 10 categories, with 7,000 images per category. The dataset has 60,000 training images and 10,000 testing images. Fashion-MNIST shares the same image size, data format and the structure of training and testing splits as MNIST. The dataset is based on the assortment on Zalando's website demonstrating different aspects of the fashion product, including front and back look, details, look with models and in an outfit.

### IV. RESULS AND DISCUSSION

#### A. Evaluation on MNIST dataset

For MNIST dataset, we compared our proposed SIFT descriptor CNN model results with different types of pooling operations experimented by Williams et al. [4]. We didn't perform data augmentation as many of the state-of-the-art models have done. This was deliberately done mainly to see the impact of different variants of pooling and SIFT descriptor alone. We take the model (refer Fig.2) as proposed by Williams et al. [4] as our baseline model. The last convolutional layer with kernel 1x1 effectively acts as a fully connected (FC) layer. The SIFT Descriptor variant of baseline model is shown in Fig. 3, where pooling layers have been avoided to achieve final convolution output size to be atleast 16x16, thus satisfying patch size constraint (p ≥ 16). The SIFT descriptor layer will give 128 feature vector for the patch size 16x16. The hybrid max-SIFT architecture concatenates the pre-softmax layer features and the concatenated feature is fed to the softmax layer. The performance of different variants is discussed in TABLE. I. The SIFT and hybrid max-SIFT variants perform better than the traditional pooling approaches. For SIFT variant, F1 score and precision are 0.996 and 0.996 respectively.

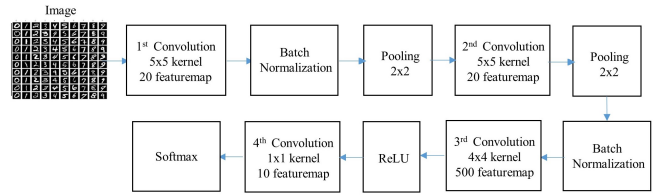

Fig. 2. Baseline CNN model [4] for MNIST dataset.

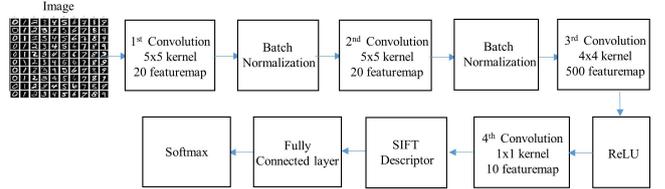

Fig. 3. Proposed SIFT descriptor CNN model for MNIST dataset.

TABLE I.  ACCURACY COMPARISON ON MNIST DATASET

| Method | MNIST |
|---|---|
| Max-Pooling [4] | 98.72 |
| Average Pooling [4] | 98.80 |
| Mixed Pooling [4] | 98.86 |
| Stochastic Pooling [4] | 98.90 |
| Wavelet Pooling [4] | 99.01 |
| SIFT Descriptor | **99.56** |
| Hybrid max-SIFT | **99.58** |

#### B. Evaluation on fashionMNIST dataset

For fashionMNIST dataset, we evaluated the performance of different variants of pooling and the proposed SIFT descriptor CNN model against the same model architecture except the pooling layer. Fig.4 shows clearly the difference between the traditional CNN and SIFT descriptor CNN architecture for the experiments on fashionMNIST dataset. The CNN architecture used has two convolutional layers with kernel shape 3x3. Dropout of 0.5 is applied at the Fully Connected (FC) layer of size 128. Rectified Linear Unit (ReLU) activation function is used in the convolutional layers and the FC layer. The SIFT descriptor layer will give 128 feature vector for the patch size 24x24. In the hybrid max-SIFT model, pooling layer and SIFT descriptor layer features are concatenated and fed to the fully connected layer of size 256. Finally the output of fully connected layer is fed to the softmax layer. For SIFT variant, F1 score and precision are 0.935 and 0.937 respectively.

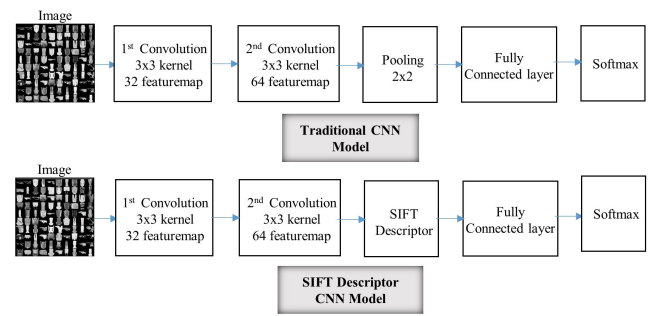

Fig. 4. Model artichecture for traditional and the proposed SIFT Desscriptor CNN Model for fashionMNIST dataset.

The remaining layers have been kept same as the traditional CNN model to make a fair comparison of the effect of SIFT descriptor layer. The performance of different variants of pooling layer is captured in TABLE. II. The SIFT and hybrid max-SIFT variants outperform the traditional pooling (max, avg, mix) approaches. In addition, the proposed SIFT-CNN model has around 8% less trainable model parameters than the traditional CNN.

TABLE II. ACCURACY COMPARISON ON FASHIONMNIST DATASET

| Method | fashionMNIST |
|---|---|
| Max-Pooling | 93.40 |
| Average Pooling | 93.15 |
| Mixed Pooling | 93.27 |
| SIFT Descriptor | **93.52** |
| Hybrid max-SIFT | **93.47** |

### C. t-SNE visualisation of pre-FC layer features

As a qualitative experiment, we visualize 2D projections of randomly selected 300 points from each class using the test set in both the FashionMNIST and MNIST datasets and compare the t-SNE [21] visualizations for the traditional and the proposed networks. We see that latter produces cleaner clusters than former, with more distinct clusters being formed on both datasets.

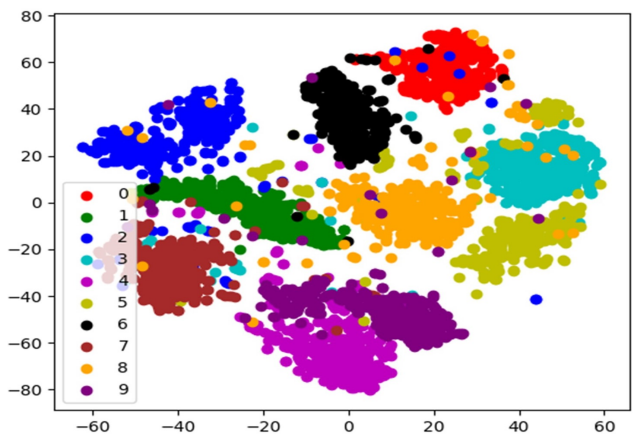

Fig. 5. Visualization for traditional CNN features on MNIST dataset. (Best viewed in color)

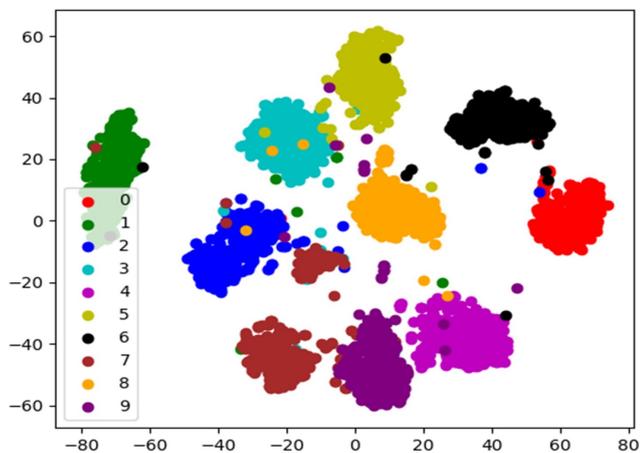

Fig. 6. Visualization for SIFT Descripor CNN features on MNIST dataset. (Best viewed in color)

For MNIST dataset, Fig.5 and Fig.6 show the visualization of the pre-fully connected (FC) layer features for the traditional and the proposed model respectively. The SIFT descriptor correctly separates different label clusters to make them visually identifiable.

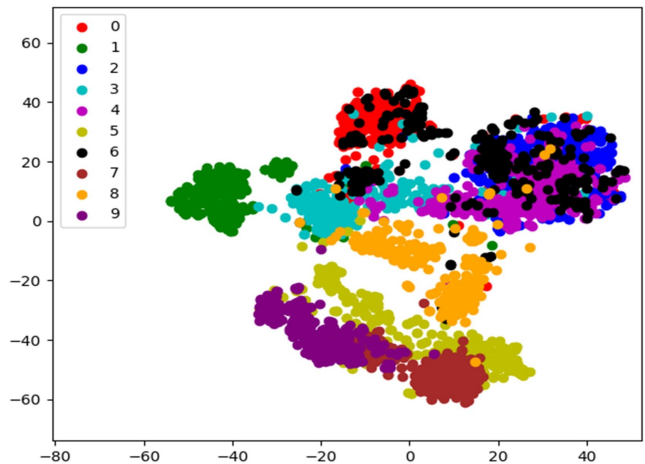

Fig. 7. Visualization for traditional CNN features on fashionMNIST dataset. (Best viewed in color)

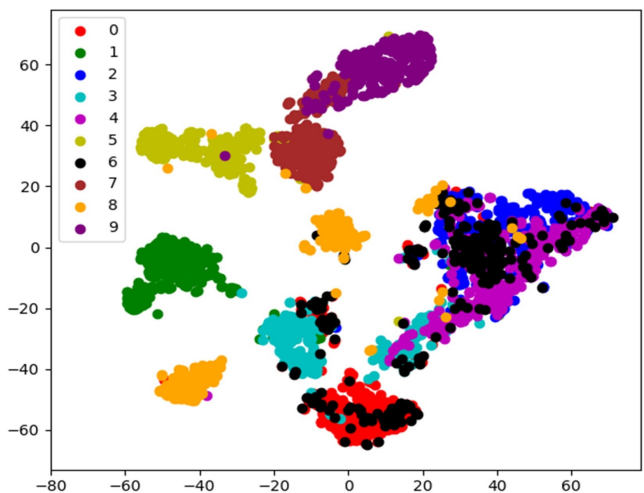

Fig. 8. Visualization for SIFT Descripor CNN features on fashionMNIST dataset. (Best viewed in color)

For fashionMNIST dataset, 2D visualization of the pre-fully connected (FC) layer features for the traditional and the proposed model is shown in Fig.7 and Fig.8 respectively. SIFT Descriptor produces more distictive clusters.

### D. Discussion

The most important drawbacks of CNN is that they are not able to model spatial relationships that well. Traditional CNN is unable to learn the internal representation of the geometrical constraints of the data, the only knowledge we have is coming from the data itself. Suppose we want to be able to detect cars in many viewpoints, we need to have these different viewpoint cars in the training set, because we did not encode the prior knowledge of the geometrical relationship into the network. Max pooling loses valuable information and also does not encode relative spatial relationships between features. Convolutional layer is good at detecting the low-level features (edges, and curves), and the high-level features (geometric shapes). But the subsequent pooling layer discards the part and whole relationships among the features. Replacing pooling layer with SIFT descriptor has captured more informative high-level features out of the convolutional output. The SIFT descriptor output is invariant to rotation, pose, illumination and partially invariant to affine distortion.

## V. CONCLUSION AND FUTURE WORK

We establish that SIFT descriptor can potentially replace traditional pooling methods used in the CNN literature. The proposed method outperforms state-of-the-art performance on MNIST and fashionMNIST datasets, when compared to other pooling approaches. SIFT descriptor provides rotation invariance, which eliminates the need of data augmentation, hence reduces training time and overfitting. Furthermore, our method has negligible computational overhead and no additional hyper-parameters to tune, thus can be easily swapped into any existing convolutional network architecture. Currently we have heuristically decided the patch size for the SIFT descriptor for experimentation on both datasets. However, finding the optimal patch size for different datasets is one possible challenge and its impact needs to be analyzed. Another possible challenge is related to images with complex background as background keypoints may affect the robustness of SIFT descriptor. For further research, novelty of the proposed method can be evaluated for different model architectures and datasets like IMAGENET and biometrics. Additionally, we plan to investigate the effect of different patch sizes on accuracy and learn the optimal value of patch size for SIFT descriptors.

## VI. ACKNOWLEDGMENT

The authors would like to acknowledge the support of Samsung R&D Institute-India, Bangalore in this work.

## VII. REFERENCES

[1] Hinton, Geoffrey E., Sara Sabour, and Nicholas Frosst. "Matrix capsules with EM routing." (2018).

[2] Sabour, Sara, Nicholas Frosst, and Geoffrey E. Hinton. "Dynamic routing between capsules." In Advances in Neural Information Processing Systems, pp. 3856-3866. 2017.

[3] Lowe, David G. "Distinctive image features from scale-invariant keypoints." International journal of computer vision 60, no. 2 (2004): 91-110.

[4] Williams, Travis, and Robert Li. "Wavelet Pooling for Convolutional Neural Networks." International Conference on Learning Representations (2018).

[5] Tolias, Giorgos, Ronan Sicre, and Hervé Jégou. "Particular object retrieval with integral max-pooling of CNN activations." arXiv preprint arXiv:1511.05879 (2015).

[6] Mishkin, Dmytro, and Jiri Matas. "All you need is a good init." *arXiv preprint arXiv:1511.06422* (2015).

[7] Lee, Chen-Yu, Patrick W. Gallagher, and Zhuowen Tu. "Generalizing pooling functions in convolutional neural networks: Mixed, gated, and tree." In Artificial Intelligence and Statistics, pp. 464-472. 2016.

[8] Nielsen, Michael A. *Neural networks and deep learning.* Determination Press, 2015.

[9] Boureau, Y-Lan, Jean Ponce, and Yann LeCun. "A theoretical analysis of feature pooling in visual recognition." In Proceedings of the 27th international conference on machine learning (ICML-10), pp. 111-118. 2010.

[10] Zeiler, Matthew D., and Rob Fergus. "Stochastic pooling for regularization of deep convolutional neural networks." arXiv preprint arXiv:1301.3557 (2013).

[11] Nitish Srivastava, Geoffrey Hinton, Alex Krizhevsky, Ilya Sutskever, and Ruslan Salakhutdinov. Dropout: A Simple Way to Prevent Neural Networks from Overfitting. Journal of Machine Learning Research, 15:1929–1958, 2014.

[12] Wan, Li, Matthew Zeiler, Sixin Zhang, Yann Le Cun, and Rob Fergus. "Regularization of neural networks using dropconnect." In *International Conference on Machine Learning*, pp. 1058-1066. 2013.

[13] Yu, Dingjun, Hanli Wang, Peiqiu Chen, and Zhihua Wei. "Mixed pooling for convolutional neural networks." In International Conference on Rough Sets and Knowledge Technology, pp. 364-375. Springer, Cham, 2014.

[14] Graham, Benjamin. "Fractional max-pooling." arXiv preprint arXiv:1412.6071 (2014).

[15] Jarrett, Kevin, Koray Kavukcuoglu, and Yann LeCun. "What is the best multi-stage architecture for object recognition?." In Computer Vision, 2009 IEEE 12th International Conference on, pp. 2146-2153. IEEE, 2009.

[16] Krizhevsky, Alex, Ilya Sutskever, and Geoffrey E. Hinton. "Imagenet classification with deep convolutional neural networks." In Advances in neural information processing systems, pp. 1097-1105. 2012.

[17] Ioffe, Sergey, and Christian Szegedy. "Batch normalization: Accelerating deep network training by reducing internal covariate shift." arXiv preprint arXiv:1502.03167 (2015).

[18] Simonyan, Karen, and Andrew Zisserman. "Very deep convolutional networks for large-scale image recognition." arXiv preprint arXiv:1409.1556 (2014).

[19] Glorot, Xavier, and Yoshua Bengio. "Understanding the difficulty of training deep feedforward neural networks." In Proceedings of the thirteenth international conference on artificial intelligence and statistics, pp. 249-256. 2010.

[20] He, Kaiming, Xiangyu Zhang, Shaoqing Ren, and Jian Sun. "Deep residual learning for image recognition." In Proceedings of the IEEE conference on computer vision and pattern recognition, pp. 770-778. 2016.

[21] Maaten, Laurens van der, and Geoffrey Hinton. "Visualizing data using t-SNE." Journal of machine learning research 9, no. Nov (2008): 2579-2605.

[22] Ding, Jun, Bo Chen, Hongwei Liu, and Mengyuan Huang. "Convolutional neural network with data augmentation for SAR target recognition." IEEE Geoscience and remote sensing letters 13, no. 3 (2016): 364-368.

[23] Deng, Li. "The MNIST database of handwritten digit images for machine learning research [best of the web]." IEEE Signal Processing Magazine 29, no. 6 (2012): 141-142.

[24] Xiao, Han, Kashif Rasul, and Roland Vollgraf. "Fashion-mnist: a novel image dataset for benchmarking machine learning algorithms." arXiv preprint arXiv:1708.07747 (2017).